\documentclass[11pt,a4paper,x11names]{article}
\usepackage{setspace}
\usepackage[acceptedWithA]{misc/tacl2021v1}
\usepackage{times}
\usepackage{xcolor}
\usepackage{graphicx}
\usepackage{xfrac} %
\usepackage[T1]{fontenc}
\usepackage[utf8]{inputenc}
\usepackage{microtype}
\usepackage{booktabs}
\usepackage{todonotes}
\usepackage{amsmath,amssymb,bbm}
\usepackage{cleveref}
\usepackage{xspace}
\usepackage{listings,algorithm, algorithmicx, algpseudocode}
\usepackage{soul} %
\usepackage[hypcap=false]{caption}
\usepackage{multirow,multicol}
\usepackage{ulem}
\usepackage{float}
\usepackage{array}
\usepackage{bm}
\usepackage{fancyvrb} %
\usepackage{xfrac}
\usepackage[shortlabels]{enumitem}
\usepackage{dsfont} %
\usepackage{url} %
\usepackage{tcolorbox}
\usepackage{makecell} %
\DeclareUrlCommand\UScore{\urlstyle{rm}}
\usepackage{anyfontsize} %

\usepackage{silence}
\setlist[itemize,enumerate]{noitemsep}

\usepackage{tabularx}

\usepackage{stmaryrd}
\usepackage{trimclip}

\crefname{lstlisting}{listing}{listings}
\Crefname{lstlisting}{Listing}{Listings}
\crefname{myequation}{equations}{equations}
\Crefname{myequation}{Equations}{Equations}
\Crefname{algorithmCaption}{Algorithm}{Algorithms}
\crefname{example}{example}{examples}
\Crefname{example}{Example}{Examples}
\crefname{prompt}{prompt}{prompts}
\Crefname{prompt}{Prompt}{Prompts}
\DeclareCaptionType{algorithmCaption}[Algorithm][List of algorithms]
\DeclareCaptionType{example}[Example][List of examples]
\DeclareCaptionType{prompt}[Prompt][List of prompts]
\DeclareCaptionType{myequation}[Equation][List of equations]

\usepackage{courier}
\usepackage{marginnote}

\definecolor{TodoColor}{rgb}{1,0.7,0.6}

\definecolor{FindingsColor}{gray}{0.85}

\makeatletter\def\Hy@Warning#1{}\makeatother
\let\svthefootnote\thefootnote
\newcommand\blankfootnote[1]{%
  \let\thefootnote\relax\footnotetext{#1}%
  \let\thefootnote\svthefootnote%
}

\newcommand{\legend}[3]{
\null\hspace{0mm}
\makebox[25mm][l]{
    \textcolor[HTML]{#1}{
    \rule[3pt]{10pt}{1.5pt}
    \hspace{-13pt}
    \raisebox{0.5pt}{\scalebox{1.5}{$\bullet$}}}
    #2
}
\makebox[8mm][l]{#3}
}

\newcommand{\legendShort}[2]{
\null\hspace{1mm}
\makebox[21mm][l]{
    \textcolor[HTML]{#1}{
    \rule[3pt]{10pt}{1.5pt}
    \hspace{-13pt}
    \raisebox{0.5pt}{\scalebox{1.5}{$\bullet$}}}
    #2
}
}

\author{%
    Vilém Zouhar
    \qquad
    Peng Cui
    \qquad
    Mrinmaya Sachan
    \AND
    \null\\[-1cm]
    \normalfont 
    Department of Computer Science, ETH Zurich \\
    \small\tt \{\href{mailto:vzouhar@ethz.ch}{\color{black} vzouhar}, \href{mailto:pcui@ethz.ch}{\color{black} pcui}, \href{mailto:msachan@ethz.ch}{\color{black} msachan}\}@ethz.ch
}

\title{How to Select Datapoints for Efficient Human Evaluation of NLG Models?}

\begin{document}
\maketitle

\footnotetext[0]{We release the \href{https://github.com/zouharvi/subset2evaluate}{subset2evaluate} package, trained models, and code for reproducing the results in this paper.}

\begin{abstract}
Human evaluation is the gold standard for evaluating text generation models.
However, it is expensive. In order to fit budgetary constraints, a random subset of the test data is often chosen in practice for human evaluation.
However, randomly selected data may not accurately represent test performance, making this approach economically inefficient for model comparison.
Thus, in this work, we develop and analyze a suite of selectors to get the most informative datapoints for human evaluation, taking the evaluation costs into account.
We show that selectors based on variance in automated metric scores, diversity in model outputs, or Item Response Theory outperform random selection. 
We further develop an approach to distill these selectors to the scenario where the model outputs are not yet available.
In particular, we introduce source-based estimators, which predict item usefulness for human evaluation just based on the source texts.
We demonstrate the efficacy of our selectors in two common NLG tasks, machine translation and summarization, and show that only $\sim$70\% of the test data is needed to produce the same evaluation result as the entire data.
\end{abstract}

\section{Introduction}

Robust model evaluation drives natural language generation (NLG) research. 
Despite improvements in automated evaluation metrics, human evaluation remains the gold standard in NLG \citep{zhou-etal-2022-deconstructing,freitag-etal-2023-results}.
Distinguishing between high-quality models requires increasingly more expert human annotations to determine which model performs better.
For example, WMT shared tasks \citep[Conference on Machine Translation,][]{kocmi2024wmt}, annually evaluate dozens of state-of-the-art machine translation models.

The dominant approach in human evaluation under budgetary constraints is to \textit{randomly} select evaluation items in the test data, as shown by the survey of \citet{ruan2024betterrandomreliablenlg}.
The random selection is clearly suboptimal because it can lead to selecting overlapping items while omitting items that impact the evaluation outcome (e.g. model ranking) more.

Thus, in this paper, we develop approaches to select a subset of test items to reduce human-evaluation cost without sacrificing evaluation accuracy.
In particular, we are interested in selecting a subset of test items that would lead to a human ranking of multiple NLG models similar to that on the entire test set.
We frame this as a subset selection problem (dubbed output-based selection, \Cref{fig:highlevel_subset_selection_outputbased}) where we are given a set of items $\mathcal{X}$ and model outputs $\mathcal{M}$, and would like to select a subset $\mathcal{Y} \subseteq \mathcal{X}$ such that the ranking of models with human evaluation on $\mathcal{Y}$ is the same as on $\mathcal{X}$.
Although automated metrics can be noisy and may not always align with human judgments, they still help identify test items that are informative for evaluation.
We build multiple data selectors that leverage these metrics, such as prioritizing challenging items, items leading to diverse model outputs, or using Item Response Theory.

However, in some evaluation scenarios, such as when organizing a large shared task, we may not yet know which models will be evaluated by our test set, or obtaining all model outputs on the whole $\mathcal{X}$ may be computationally infeasible due to the size of $\mathcal{X}$.
In this setting, called source-based selection (\Cref{fig:highlevel_subset_selection_sourcebased}), the standard methods for output-based selection cannot be used.
However, by distilling the output-based selection methods, we build predictors that only use the item input to predict the expected item difficulty or likelihood that it leads to diverse model outputs.

We demonstrate the efficacy of our data selection approach with case studies on two typical natural language generation tasks: machine translation and summarization. 
Our key contributions include:
\begin{itemize}[noitemsep,left=0mm]
\item framing the task of informative subset selection for evaluation in two variants,
\item multiple evaluation subset selection methods, including cost- and document-aware selection,
\item selector distillation for source-based selection,
\item package \href{https://github.com/zouharvi/subset2evaluate}{subset2evaluate} for budget-efficient test set construction for model evaluation.
\end{itemize}

\noindent
This paper is structured as follows.
We formalize the problem in \Cref{sec:problem_statement}, describe the output-based and source-based selectors in \Cref{sec:outputbased_selectors,sec:sourcebased_selectors}. 
We show our results on efficient machine translation evaluation in \Cref{sec:results_mt} and our results on evaluating summarization models
in \Cref{sec:results_summ}.
In particular, we find that in the annual WMT evaluation for machine translation models, our methods yield the same evaluation result (model ranking) as random sampling, but with only 70\% of human annotations.
Given the high cost of human evaluation, this is a non-trivial cost saving.

\section{Problem Statement}
\label{sec:problem_statement}

\begin{figure}[t]
\centering
\includegraphics[width=0.95\linewidth]{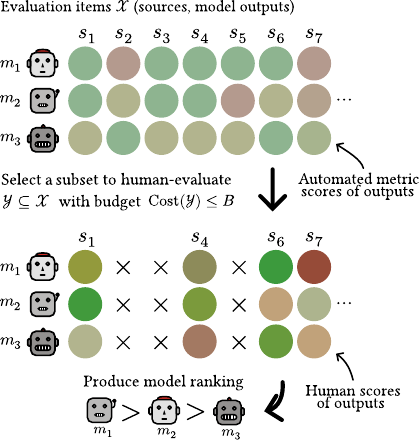}

\caption{
Output-based informative subset selection approach.
Given model outputs and automated metrics, we select items to be human-evaluated on which the final model ranking can be computed.
}
\label{fig:highlevel_subset_selection_outputbased}

\end{figure}

We are given a set of items $\mathcal{X}$ and a set of models $\mathcal{M}$ that we wish to rank according to $\mathcal{X}$.
Here, each item $x \in \mathcal{X}$ is an input text and $m(x)$ is the output of the model $m$ in the NLG task.
For example, in machine translation, each $x$ is the input in the source language and $m(x)$ is the corresponding translation.
Since $|\mathcal{X}|$ is very large (exceeding the human evaluation budget $B$), we seek a subset $\mathcal{Y} \subseteq \mathcal{X}$ such that the ranking of models on $\mathcal{Y}$ can be as close to that on $\mathcal{X}$ as possible.
An illustration of the problem is also shown in \Cref{fig:highlevel_subset_selection_outputbased}.

\paragraph{Human evaluation set selection.}
In order to obtain
a subset $\mathcal{Y} \subseteq \mathcal{X}$ to be human-evaluated,
we quantify the cost of human evaluations on $\mathcal{Y}$ with $\mathrm{Cost}(\mathcal{Y})$ and the usefulness of the subset for evaluation with $\mathrm{Utility}(\mathcal{Y})$.
In the ideal case, the utility of the set $\mathcal{Y}$ indicates how close the human ranking of models on items $y \in \mathcal{Y}$ is to the human ranking of models on whole $\mathcal{X}$.
We frame this as a subset selection problem as follows:
\begin{align}
\underset{\mathcal{Y}\subseteq\mathcal{X}}{\arg\max} \,\, \mathrm{Utility}(\mathcal{Y}) \quad
\text{s.t.} \quad
\mathrm{Cost}(\mathcal{Y}) \leq B \label{eq:problem_formulation}
\end{align}

\noindent
In our work, we make two simplifications.
First, we note that the cost of evaluating a set of items is the sum of costs for evaluating individual items, i.e. $\mathrm{Cost}(\mathcal{Y}) = \sum_{y\in \mathcal{Y}} \mathrm{Cost}(y)$.
This assumption is generally true if the human evaluation of items $y$ is carried out one by one.
Second, we assume that the utility of the set $\mathcal{Y}$ is the sum of the utilities of items $y \in \mathcal{Y}$.
Note that this is generally not true, as similar datapoints may not offer an extra marginal utility for model ranking. However, in our experiments, we find that this assumption leads to good empirical results.
Moreover, our preliminary experiments on modeling diversity of datapoints did not lead to better performance (see methods without this assumption in \Cref{sec:other_methods}).

With the two assumptions, we rewrite \Cref{eq:problem_formulation} using 0/1 indicator variables $z_x$, which becomes a 0-1 knapsack problem.
\begin{align}
\text{maximize}\quad & \sum_{x \in \mathcal{X}} z_x \cdot \mathrm{Utility}(x) \nonumber \\
\text{subject to}\quad & \sum_{x \in \mathcal{X}} z_x \cdot \mathrm{Cost}(x) \leq B \nonumber \\
&\mathrm{and}\, \forall x \in \mathcal{X}: z_x \in \{0, 1\} \label{eq:problem_formulation_knapsack}
\end{align}
This problem is generally NP-complete \citep{karp1975reducibility}.
However, for constant $\mathrm{Cost}(.)$ and positive item utilities, we can find an optimal solution taking the items with the highest utilities until the budget $B$ is reached.
For a non-constant $\mathrm{Cost}(.)$, we find an approximate solution using integer linear programming \citep{huangfu2018parallelizing}.

Next in \Cref{sec:outputbased_selectors}, we present multiple ways to approximate the utility of an item $\mathrm{Utility}(x)$.

\subsection{Subset selection evaluation}
\label{sec:subset_selection_evaluation}
Once we select a subset of data for human evaluation, how do we know if it is good?
We evaluate the quality of the selected subset with \textbf{soft pairwise accuracy} \citep{thompson-etal-2024-improving} between the model ranking on the subset and the model ranking on the entire set.
The accuracy reveals how close our evaluation result (model ranking) is to the true evaluation result (model ranking on the whole set).
Note that this meta-evaluation requires the presence of human scores.

There are many ways to evaluate the similarity of model rankings \citep{deutsch2021statistical,deutsch-etal-2023-ties}.
In contrast to Spearman \citep{spearman} or Kendall \citep{kendall} correlations, soft pairwise accuracy inherently also accounts for the statistical power of the ranking.
We further discuss the choice of this meta-evaluation in \Cref{sec:other_measures}.

\subsection{Source- vs. Output-based selection}

In many practical scenarios, for example, in the WMT shared tasks, the size of $\mathcal{X}$ may be so large that it is too expensive to even obtain model outputs on the entire dataset for all models.
Furthermore, we may not yet know all the participating models $\mathcal{M}$ in advance, for example, when preparing blind test sets for a shared task.
For such scenarios, we introduce source-based data selection.
In contrast to output-based data selection, which models item utilities with the knowledge of model outputs $m(x)$ for $m \in \mathcal{M}$, source-based data selection models $\mathrm{Utility}(x)$ based solely on the input $x$.



\section{Output-based Selectors}
\label{sec:outputbased_selectors}

We begin by describing various utility functions for output-based subset selection to use in \Cref{eq:problem_formulation}.
We will describe methods for source-based selection later, as these rely on the methods for output-based selection.
Finally, we will compare data selection approaches using the defined utilities with a random subset selection baseline that randomly chooses a subset of data with the cost $B$ for human evaluation.
We run this random sampling 100 times and compute the corresponding confidence intervals with Student's t distribution.



\subsection{Metrics moment}

We first introduce two utility functions for selecting informative items based on the distribution moments induced by automated metric scores.
Metrics provide us with a coarse estimate of item difficulty, and the shape of the distribution can be used to select high-impact items for human evaluation.

Our first heuristic for defining utility is based on average metric scores.
If, on average, an item receives a low metric score across multiple model outputs, then it can be perceived as being difficult \citep{don-yehiya-etal-2022-prequel}.
Thus, average metric scores correlate negatively with the difficulty of the item for NLG models.
Based on this, the first heuristic selects items with highest difficulty, i.e., lowest average metric score:
\begin{align}
&\hspace{-3mm}\textsc{MetricAvg}(x, \mathcal{M}) =
\\
&\quad\qquad\qquad\qquad
-\sum_{m \in \mathcal{M}} \frac{\mathrm{metric}(x, m(x))}{|\mathcal{M}|}  \nonumber
\end{align}

\noindent
However, items where all models produce very high quality or very low quality outputs do not contribute much to the final model ranking.
Thus, to prioritize high-impact items that highlight differences between models, our second metric measures variance in metrics across models, which impact the final model ranking the most:
\begin{align}
&\hspace{-0.5mm}\textsc{MetricVar}(x, \mathcal{M}) = \\ \nonumber
&\sum_{m \in \mathcal{M}} \frac{\left( \mathrm{metric}(x, m(x)){-}\textsc{MetricAvg}(x, \mathcal{M}) \right)^2}{|\mathcal{M}|}
\end{align}


\subsection{Metric consistency}
Another approach to make use of the shape of the distribution of metric scores over various models is to consider items where the automated metrics show model ranking consistent with the whole set $\mathcal{X}$.
Therefore, if an item is predictive of the overall model ranking based on metrics, it might also be predictive of the overall model ranking based on human scores:
\begin{align}
& \hspace{-5mm}\textsc{MetricCons}(x, \mathcal{M}) = \\
& \mathrm{RankCorr} \Bigg(
\substack{\langle\mathrm{metric}(x, m(x))|m \in \mathcal{M} \rangle,\hspace{11mm} \\[0.5em] \langle\sum_{x'\in \mathcal{X}} \mathrm{metric}(x', m(x'))|m \in \mathcal{M} \rangle} \Bigg) \nonumber
\end{align}
The ranking correlation might be based on Kendall, pairwise accuracy, or Spearman, which we use.
This method relates to greedy coreset construction, which we discuss in \Cref{sec:related_work} and \Cref{sec:other_methods}.

\subsection{Output diversity}
While metrics variance prioritizes items that lead to outputs of different qualities,
our fourth metric goes a step further and prioritizes items that lead to diverse \textit{outputs} as evaluating identical outputs from different models is not useful:
\begin{align}
&\hspace{-6mm}\textsc{Diversity}(x, \mathcal{M}) =\\ \nonumber
& -\frac{\sum_{m_1, m_2 \in \mathcal{M}} \mathrm{sim}(m_1(x), m_2(x))}{|\mathcal{M}|^2}
\label{eq:diversity}
\end{align}
Output diversity can be captured by average text similarity among the outputs, for example with pariwise  unigram overlap, text-matching metrics chrF \citep{popovic-2015-chrf} or BLEU \citep{papineni-etal-2002-bleu}, or embedding similarity.
We primarily use embedding similarity with details in \Cref{sec:implementation}, but also evaluate other similarity metrics in \Cref{sec:other_methods}.
We take the negative value of the average to prioritize items with a lower similarity across outputs.

\begin{figure}[t]
\centering
\includegraphics[width=0.95\linewidth]{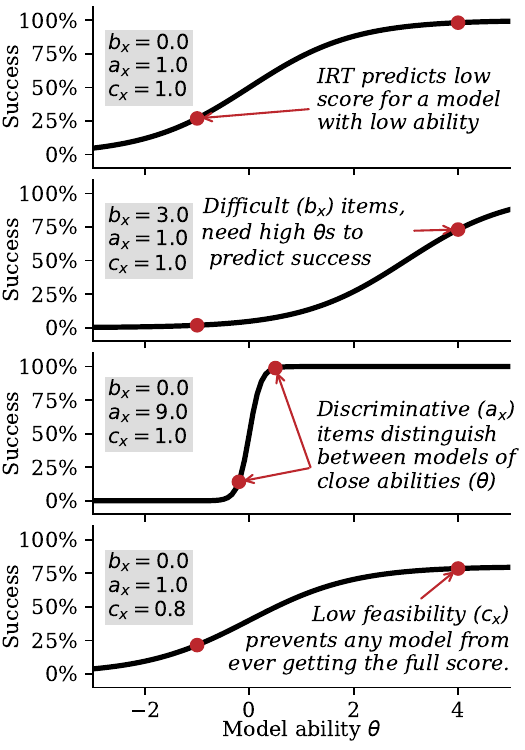}

\vspace{-3mm}
\caption{Illustration on how item response theory predicts model success. The parameters are difficulty $b_s$ (shift on x-axis), discriminability $a_s$ (slope), and feasibility $c_s$ (upper bound).
}
\label{fig:40-mock_irt}

\vspace{-1mm}
\end{figure}

\subsection{Item response theory}


Previous works on informative test construction often use Item Response Theory (IRT, \citealp{santor1998progress}), which is a principled approach to the problem.
Inspired by psychometrics and educational sciences, IRT provides a way to create a budget-efficient test for evaluating and comparing various students.
Given a set of students' responses to a set of test items, IRT models the probability that a student $m$ with a given ability level $\theta_m$ will answer a question $x$ correctly as the standard logistic function:
\begin{align}
    \hspace{-2.5mm}
    p(r_{m,x}{=}1) = \frac{c_x}{1+\mathrm{exp}[-a_x(\theta_m-b_x)]},
    \hspace{-1mm}
\end{align}

Here, the student response ($r$) is usually a binary label (correct or incorrect), 
$a_x$ denotes the \textbf{discriminability} of the item, $b_x$ denotes the \textbf{difficulty} of the item $x$, and $c_x$ denotes the maximum achievable chance of answering the items (e.g. due to ambiguity of the item).
The item parameters define the shape of the standard logistic function: for items with high discriminability $a_x$, even very close students can be distinguished because a small change in $\theta_m$ changes the prediction.

We draw on this analogy and fit an IRT model to predict the metric scores $\mathrm{metric}(x, m(x))$ for various models $m$ on the items $x$.
However, in NLG, the metric scores are usually not binary, but continuous.
Some approaches bypass this by binarizing the continuous variable:
$r_{m,x} := \mathbbm{1}[\mathrm{metric}(x, m(x)) > 0.5]$
\citep{polo2024tinybenchmarks}.
However, this leads to a loss of information, and we can redefine the objective to directly predicting the continuous score $r_{m,x} :=\mathrm{metric}(x, m(x))$ as in other works in psychometry \citep{noel2007beta}:
\begin{align}
    \hat r_{m,x} = \frac{c_x}{1+\mathrm{exp}[-a_x(\theta_m-b_x)]}, \label{eq:irt}
\end{align}

See an illustration of IRT predictions and parameters $b_x$, $a_x$, and $c_x$ in \Cref{fig:40-mock_irt}.
Following standard practice, we optimize the IRT model
with stochastic variational inference \citep{wu2020variationalitemresponsetheory,rodriguez-etal-2021-evaluation,py-irt} using the ELBO loss.
In practice, this corresponds to making the estimated response $\hat r_{m,x}$ be closer to the true response $r_{m,x}$.

We repurpose this method from evaluating students to evaluating models and use both the item discriminability $a_x$, and difficulty $b_x$ for utility:
\begin{align}
\textsc{DiffDisc}\, (x) = a_x \times b_x
\end{align}
We use a product instead of addition because $a_x$ and $b_x$ may have different scales but consider also other formulas in \Cref{sec:other_methods}.





\section{Source-Based Selectors}
\label{sec:sourcebased_selectors}
\begin{figure}[t]
\centering
\includegraphics[width=0.95\linewidth]{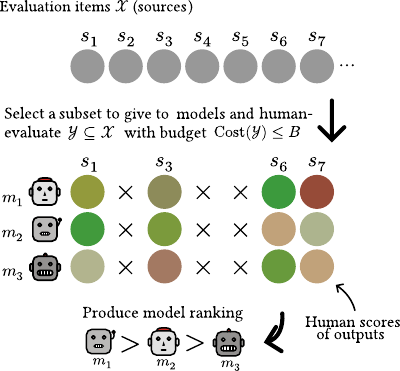}
\caption{
Source-based variant of subset selection (in contrast with output-based in \Cref{fig:highlevel_subset_selection_outputbased}).
Given just the item inputs, we select items to be given to models and their outputs to be human-evaluated.
}
\label{fig:highlevel_subset_selection_sourcebased}

\vspace{-2mm}
\end{figure}

Next, we discuss approaches for source-based data selection
where the model outputs and metric scores are not available yet.
This is illustrated in \Cref{fig:highlevel_subset_selection_sourcebased}.
We provide two ways to adapt our output-based methods to the case where we select datapoints given only source texts: (1) estimating item utility just based on $x$ via {\bf model distillation}, and (2) creating an {\bf artificial crowd} of models imitating $\mathcal{M}$.

\subsection{Item utility distillation}

Calculating item utilities in output-based selection requires access to the outputs of models $\mathcal{M}$.
To circumvent this requirement in source-based selection, we develop a model distillation approach to fit a model that predicts the item utility in the absence of $\mathcal{M}$, just based on the item input $x$:
\begin{align}
\mathrm{Utility}^\mathrm{src}(x) \xrightarrow{\text{train}} \mathrm{Utility}(x, \mathcal{M})
\label{eq:distillation}
\end{align}
We use the architecture of a learned NLG metric \citep{rei-etal-2020-comet} to predict the utilities.
The input text $x$ is encoded with a pre-trained language model, and a regression head is trained with MSE loss predicting output-based utilities.
As a result, we obtain new source-based item utility estimators: \textsc{MetricAvg}\textsuperscript{src}, \textsc{MetricVar}\textsuperscript{src}, \textsc{MetricCons}\textsuperscript{src}, \textsc{MetricDiversity}\textsuperscript{src}, and \textsc{DiffDisc}\textsuperscript{src}.
For IRT, \citet{benedetto2020r2de,byrd-srivastava-2022-predicting} build functions that directly predict the item discriminability $a_x$ and difficulty $b_x$ for unseen items.
In contrast, our predictor directly predicts the product $a_x \times b_x$ used for computing the item utility.

An advantage of source-based item utilities over output-based ones is that we can train the model to predict the utilities based on human scores as opposed to metric scores, if these exist in the training data.
In the context of machine translation, similar models are used to estimate the properties of translations and sources \citep{rei-etal-2020-comet,don-yehiya-etal-2022-prequel,zouhar-etal-2023-poor,perrella-etal-2024-guardians}.

See implementation details in \Cref{sec:implementation}.
Note that this method only works on novel items that are similar to those in the training data.

\begin{figure}[t]

\centering
\begin{minipage}{\linewidth}
\input{img/13-main_outputbased_spa_legend.tex}
\end{minipage}

\vspace{-1mm}

\includegraphics[width=\linewidth]{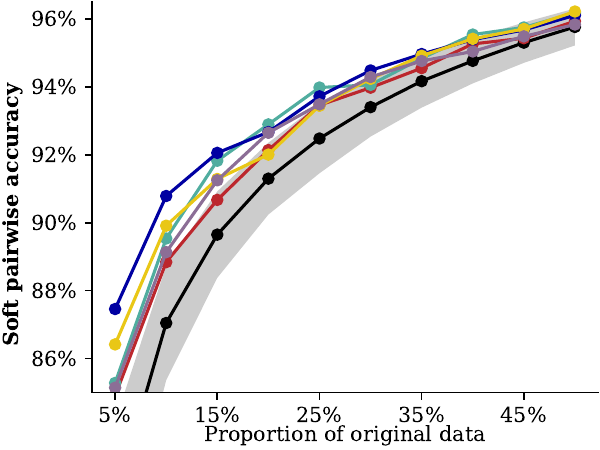}

\caption{
Main \textbf{output-based results for machine translation} (WMT23) with soft pairwise accuracy.
Subset selection methods are based either on MetricX-23 or output diversity.
We also show 90\% t-distribution confidence intervals for the Random Selector from 100 runs.
Numbers in the legend show average soft pairwise accuracy across all data proportions.
}
\label{fig:main_outputbased}

\vspace{-2mm}
\end{figure}

\subsection{Artificial crowd}

In some cases, we might not know the complete set of models $\mathcal{M}$ or not have the computational capacity to compute the output for all the models in $\mathcal{M}$.
However, we might still have some prior knowledge of $\mathcal{M}$, such as knowing which particular language models will be present.
We now assume that we know $\mathcal{M}'$ such that $\mathcal{M}' \approx \mathcal{M}$ in some capacity.
Then we can use the outputs from the models $\mathcal{M}'$ to approximate the output-based utilities without exact knowledge of $\mathcal{M}$:
\begin{align}
\textsc{Utility}(x, \mathcal{M}') \approx \textsc{Utility}(x, \mathcal{M})
\end{align}
In our implementation, we take a subset of the original $\mathcal{M}$, simulating the case where we are willing to compute model outputs for at least a portion of the models.
The approach of designing an artificial crowd was previously used by \citet{lalor2019learning} to train IRT in sentiment classification and natural language inference tasks.



\section{Case Study 1: Machine Translation}
\label{sec:results_mt}
As our first case study for budget-efficient human evaluation, we consider machine translation, particularly the setting in the \href{https://www2.statmt.org/wmt24/}{general WMT shared task} \citep{kocmi2024wmt}.
For WMT and similar venues, both source-based and output-based variations come into play sequentially:
(1) all source data is collected,
(2) an initial subset is created based on just sources,
(3) the initial subset is distributed to participants \& outputs are collected,
(4) automated metrics are computed,
(5) the final subset is created based on sources and outputs,
(6) the final subset is human-evaluated, and
(7) the model ranking is produced.

We first describe the experimental setup by which we evaluate our data selection methods.
We then present the results for the simpler of the two scenarios: output-based selection, considering both segment-level translation and document-level translation (\Cref{sec:mt_output}).
Then, we explore the case considering the annotation cost of individual items and their impact on subset selection (\Cref{sec:annotation_costs}).
We conclude with the results for the source-based data selection task (\Cref{sec:results_mt_sourcebased}).
In our results, we focus on selecting data subsets of sizes ranging from 5\% to 50\% of the original test set.

In order to satisfy the positivity constraints of the item utility functions (\Cref{sec:problem_statement}), we shift all the utilities to be positive by a constant.
In our initial experiments, we assume a constant human evaluation cost for each item.

\makeatletter
\DeclareRobustCommand{\shortto}{%
  \mathrel{\mathpalette\short@to\relax}%
}

\newcommand{\short@to}[2]{%
  \mkern2mu
  \clipbox{{.5\width} 0 0 0}{$\m@th#1\vphantom{+}{\shortrightarrow}$}%
  }
\makeatother
\newcommand{\tto}[2]{#1$\shortto$#2}

\paragraph{Data setup.}
For our experiments, we use the human annotation data from \href{https://github.com/google-research/mt-metrics-eval}{publicly available} past WMT campaigns.
We include only the language pairs and WMT years with at least 500 human annotations per model.
As a result, we ended with 33 campaigns with 31k source items and 395k translations.
Of these, we use the nine language pairs from WMT 2023 that contain MetricX-23 (\tto{Czech}{Ukrainian}, \tto{German}{English}, \tto{Japanese}{English}, {Chinese}$\shortto$ {English}, \tto{English}{Czech}, \tto{English}{German}, {English}$\shortto$ {Japanese}, \tto{English}{Chinese}, {Hebrew}$\shortto${English}).
Unless specified otherwise, we average the results (soft pairwise accuracy) across all languages.

\subsection{Output-based selection for WMT}
\label{sec:mt_output}

We evaluated the efficacy of our proposed methods using soft pairwise accuracy (\Cref{sec:subset_selection_evaluation}).
The results of output-based selection (that is, when models and metric scores are known) are shown in \Cref{fig:main_outputbased} at specific subset sizes.
We compare these methods with random selection, which has been reported to be a strong baseline in previous works \citep{rodriguez-etal-2021-evaluation,park2022active,polo2024tinybenchmarks}.
We discuss the reason for random being a strong baseline in \Cref{sec:other_measures}.

We find that approaches based on metrics moments (average and variance of automated metrics) surpass the random baseline confidence interval only with a slight margin when measured by soft pairwise accuracy.
Additionally, the IRT model performs slightly better than the metrics moment approaches, and diversity performs best.
Note that even improvements within the random baseline confidence interval but above the mean are meaningful, because by sampling randomly, we are likely to arrive at a subset with soft pairwise accuracy anywhere within the shaded area, so worse than the methods.
The results across individual WMT years and languages are shown in the Appendix \Cref{tab:48-everything_results_spa}.
In \Cref{sec:other_measures} we further confirm that even when the subset selection methods are meta-evaluated differently from soft pairwise accuracy (e.g. by correlations between model rankings), they still consistently outperform random selection.


\newcommand{\conf}[1]{\tiny $\pm$#1\%\hspace{-2mm}}
\newcommand{\confoffset}{\phantom{\conf{0.30}}}

\begin{table}[t]
\centering
\small
\begin{tabular}{lcc}
\toprule
\bf Method & \bf Output-based & \bf Source-based\textsuperscript{(src)}  \\
\midrule 
Random     & \confoffset 91.3\%\conf{0.30} & \confoffset 91.3\%\conf{0.30} \\
MetricAvg  & 92.4\% & 91.8\% \\
MetricVar  & 92.6\% & 91.9\% \\
MetricCons & 93.2\% & 93.0\% \\
Diversity  & 92.9\% & 92.6\% \\
DiffDisc   & 92.4\% & 92.1\% \\
\bottomrule
\end{tabular}
\caption{Subset selection with balanced domains. The top-$\sfrac{B}{|\mathcal{D}|}$ of item utilities within each domain in $\mathcal{D}$ is taken for particular budget $B$.
$\pm$ shows 90\% t-test confidence interval from 100 runs.
Results are averaged across languages and subset sizes.
}

\bigskip

\begin{tabular}{lcc}
\toprule
\bf Method & \bf Output-based & \bf Source-based\textsuperscript{(src)} \\
\midrule 
Random & \confoffset 87.2\%\conf{0.28} & \confoffset 87.2\%\conf{0.28} \\
MetricAvg & 90.5\% & 88.2\% \\
MetricVar & 89.5\% & 88.7\% \\
MetricCons & 89.3\% & 89.2\% \\
Diversity & 89.6\% & 89.0\% \\
DiffDisc & 90.1\% & 89.2\% \\
\bottomrule
\end{tabular}
\caption{Document-level subset selection. The item utilities are averaged to create document-level utilities out of which a subset is chosen.
$\pm$ shows 90\% t-test confidence interval from 100 runs.
Results are averaged across languages and subset sizes.
}
\label{tab:24-document_level}
\label{tab:25_domain_balanced}
\end{table}


\medskip

We now cover two practical aspects of dataset creation: (1) selecting with some apriori item distribution, such as by having a specific number of items in each data domain, and (2) selecting higher-level item units, such as documents, while items correspond to paragraphs within the document.

\paragraph{Balancing domains.}
Items in the WMT data come from multiple distinct domains, such as news, social, or chat.
As a result, the subset would measure fewer ability directions, which would lead to more significance when comparing models.
To test this, we enforce equal number of selected items in each domain.
The results in \Cref{tab:25_domain_balanced} (left column) show that even with this balancing, our methods perform much better than random selection.

\paragraph{Selecting document-level items.}
Recently, MT evaluation has shifted to document-level translation evaluations \citep[][inter alia]{kocmi2024wmt} where entire document translations are now evaluated.
We extend our methods to this setting by defining the utility of a document as the average of various sentence-level utilities defined in this paper.
We now consider entire documents to be items where the cost and utility of the document are the sum of costs of evaluating sentence-level translations in the document and average of sentence-level utilities, respectively. 
The results in \Cref{tab:24-document_level} (left column) mimic the main results for item-level selection and outperform random selection.

\paragraph{Other methods.}
In \Cref{sec:other_methods} we describe and evaluate other subset selection methods, spanning baselines, oracles, and related work.
Notably, apart from oracles, no method consistently outperforms the methods mentioned in the main paper.

\subsection{Accounting for human annotation cost\linebreak in subset selection}
\label{sec:annotation_costs}

The cost of human evaluation of model outputs is usually determined by the total time that human annotators spend evaluating them.
However, different annotation items take different times to annotate due to their lengths.
For example, the machine translation evaluation campaign of \citet{kocmi-etal-2024-error} contains very short 3-word items that take less than 10 seconds to annotate but also multi-sentence items that take more than 3 minutes to annotate each.

As human evaluation time is not known during the subset selection stage, we use
source length as a rough measure for human evaluation time.
\citet{kocmi-etal-2024-error} provide a human evaluation of the machine translation dataset with evaluation times.
We used this data to approximate the human evaluation time as a linear function of source length.
$\hat{t}(x) = 0.15 \cdot |x| + 33.7$.
This approximation is weakly correlated with the real human evaluation time ($\rho{=}0.24$).
Furthermore, items with the highest difficulty (lowest metric scores, MetricAvg) correlate positively ($\rho{=}0.33$) with human evaluation time.
Therefore, an item difficult for models usually has a higher annotation cost compared to the rest of the test set.
Selecting the most difficult items to human evaluate may thus be suboptimal and lead to higher costs.

To avoid accidentally increasing the cost of the evaluation, we need to take the cost into account during subset selection.
For this, we approximate the solution to the optimization in \Cref{eq:problem_formulation_knapsack} with integer linear programming.
The results in \Cref{tab:38-cost_aware_selection} show that cost-aware subset selection can lead to a very high soft pairwise accuracy even with a limited budget.

\begin{table}[t]

\centering
\small
\begin{tabular}{lcc}
\toprule
\bf Method & \bf Output-based & \bf Source-based \textsuperscript{(src)}  \\
\midrule 
Random & \,\,\,93.9\%$^{\Delta\text{2.3}\%}_{\pm0.38\%}$ & \,\,\,93.9\%$^{\Delta\text{2.3}\%}_{\pm0.38\%}$ \\[0.8em]
MetricAvg & 94.6\%\textsuperscript{$\Delta$2.1\%} & 93.7\%\textsuperscript{$\Delta$1.2\%} \\
MetricVar & 94.8\%\textsuperscript{$\Delta$1.8\%} & 93.9\%\textsuperscript{$\Delta$0.9\%} \\
MetricCons & 94.5\%\textsuperscript{$\Delta$1.2\%} & 94.3\%\textsuperscript{$\Delta$1.0\%} \\
Diversity &  94.4\%\textsuperscript{$\Delta$1.4\%} & 94.2\%\textsuperscript{$\Delta$1.2\%} \\
Diff.$\times$Disc. & 94.7\%\textsuperscript{$\Delta$1.9\%} & 94.2\%\textsuperscript{$\Delta$1.5\%} \\
\bottomrule
\end{tabular}
\caption{Results for methods with cost-aware subset selection with integer linear programming (Equation \ref{eq:problem_formulation_knapsack}). The sizes of the resulting subsets can differ but have the same cost.
$\Delta$ indicates improvements against cost-unaware selection in \Cref{fig:main_outputbased}.
$\pm$ shows 90\% t-test confidence interval from 100 runs.
Results are averaged across languages and subset sizes.
}
\label{tab:38-cost_aware_selection}
\end{table}

\begin{figure}[t]

\centering
\begin{minipage}{\linewidth}
\input{img/14-main_sourcebased_spa_legend.tex}
\end{minipage}

\vspace{-1mm}

\includegraphics[width=\linewidth]{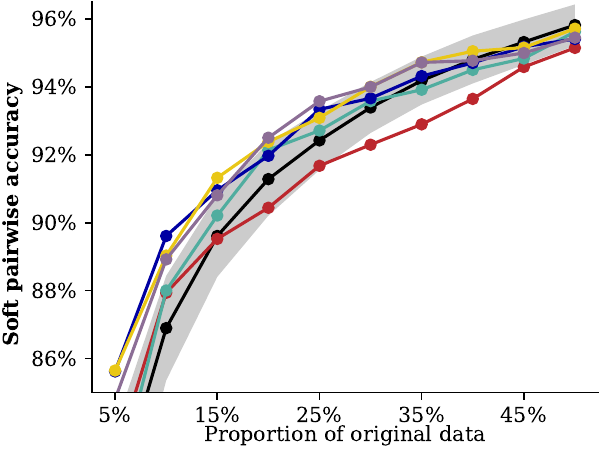}

\vspace{-1mm}

\caption{
Main \textbf{source-based results for machine translation} (WMT23) with soft pairwise accuracy.
Utility predictors are based either on distilling human scores or output diversity.
We also show 90\% t-distribution confidence intervals for the Random Selector from 100 runs.
Numbers in the legend show average soft pairwise accuracy across all data proportions.
}
\label{fig:main_sourcebased}

\vspace{-2mm}
\end{figure}


\begin{table}[t]
\centering
\small
\begin{tabular}{lc>{\hspace{-2mm}}c}
\toprule
\bf Method & \bf Artificial Crowd & \bf Source-based\textsuperscript{(src)} \\
\midrule 
Random & \confoffset 91.8\%\conf{0.89} & \confoffset 91.8\%\conf{0.89} \\
MetricAvg  & 92.0\% & 90.7\% \\
MetricVar  & 92.5\% & 92.1\% \\
MetricCons & 92.3\% & 92.1\% \\
Diversity  & 92.6\% & 92.5\% \\
DiffDisc   & 91.9\% & 92.4\% \\
\bottomrule
\end{tabular}
\caption{Output-based selection methods applied to source-based selection using artificial crowd of size of 4 models, randomly sampled from $\mathcal{M}$. The results differ from source-based selection (\Cref{fig:main_sourcebased}) because the selection is evaluated on $\mathcal{M}\setminus \mathcal{M}'$.
$\pm$ shows 90\% t-test confidence interval from 100 runs.
Averaged across languages and subset sizes.
}
\label{tab:15-artificial_crowd}
\end{table}

\subsection{Source-based selection for WMT}
\label{sec:results_mt_sourcebased}

We now describe the source-based selection case where we do not yet know the model outputs for $\mathcal{X}$.
The first approach is to distill the item utilities with a source-based predictor.

We train all item utility estimator models by distilling utilities (Equation \ref{eq:distillation}) only up to WMT22 to avoid contamination of the evaluation on WMT23.
\Cref{fig:main_sourcebased} shows the source-based selection results, which can be applied to unseen items without model outputs, unlike output-based selection.
Although the source-based estimation of metric moments performs worse than in output-based selection (\Cref{fig:main_outputbased}), it still outperforms random selection, although not outside the confidence interval.
Diversity\textsuperscript{src}, which predicts output diversity, and Diff.$\times$Disc.\textsuperscript{src}, which predict the latent IRT parameters, perform the best, though still falling short of their output-based selection versions.

For the artificial crowd approach, we assume that we know the model output on $\mathcal{X}$ for at least some models.
However, this does not require past evaluation data.
\Cref{tab:15-artificial_crowd} shows that the previous distillation methods for source-based selection perform on par with the artificial crowd approach.
The choice whether to use an artificial crowd or item utility predictors for the source-based subset selection case then depends on availability of at least some model outputs or past evaluation data on which an estimator can be trained.

\paragraph{Unfair evaluation bias.}
Using the artificial crowd might lead to a biased test set construction when used together with MetricAvg.
If a model $m$ is being evaluated ($m \in \mathcal{M}$) and is also part of the artificial crowd ($m \in \mathcal{M}'$), then $m$ is at a disadvantage because we selected difficult items for $m$ but not necessarily for $\mathcal{M} \setminus \{m\}$.
This can also happen distributionally on a higher level when we use an artificial crowd consisting of some kind of models, such as multilingual language models, which will then make the test set more difficult to multilingual language model specifically.

The same problem extends to MetricAvg\textsuperscript{src}, or any other learned metrics, when we consider that they were trained on human quality assessments of some model outputs $\mathcal{M}_\mathrm{prev}$.
Again, if a model $m$ that is being evaluated $m \in \mathcal{M}$ is also $m \in \mathcal{M}_\mathrm{prev}$, then $m$ is again at a disadvantage.

Together with the potential of selecting costly-to-evaluate items, we advise against MetricAvg for subset selection for both output- and source-based variants, especially when using an artificial crowd.





\subsection{Subset selection cuts costs substantially}


At first glance, the improvements over random selection in both scenarios appear minor.
Although previous work on active learning and subset selection agrees that random sampling is a strong baseline \citep{pmlr-v37-wei15,rodriguez-etal-2021-evaluation,park2022active,polo2024tinybenchmarks}, the improvements shown do matter, when applied at a larger scale.
To show this, we ask: \textit{What budget would we need to arrive at the same evaluation result as random sampling with budget $B$?}
The answer can be obtained with the following formula:
\begin{align}
\hat C = \mathrm{min}\, \{C | \mathrm{SPA}(\mathcal{Y}^\dagger_{\leq C}) \geq \mathrm{SPA}(\mathcal{Y}^R_{\leq B})\}
\end{align}
\noindent
where $\mathrm{SPA}$ is soft pairwise accuracy, $\mathcal{Y}^\dagger_{\leq C}$ is a subset within budget $C$ that is comparable to a random subset $\mathcal{Y}^R_{\leq B}$ of budget $B$. 

In \Cref{tab:35-evaluation_parity}, we quantify the cost-efficiency of our subset selection approach by reporting the proportion of number of datapoints in our subset that achieves the same evaluation result as random sampling, $\hat C/B$.
We find that even the simple diversity--based utility approach achieves the same or better soft pairwise accuracy with only 77\% of the data as random sampling on the machine translation evaluation task.
This means that to achieve the same human evaluation quality, we need to now pay for evaluation of only a portion of the test data.
At the scale of industrial evaluation of NLP models, which gets bigger each year, these economic implications are substantial.

\begin{table}[t]
\centering
\small
\begin{tabular}{lcc}
\toprule
\bf Method & \bf Output-based & \bf Source-based \textsuperscript{(src)}  \\
\midrule 
\sout{Random} & \sout{100.0\%}\phantom{0} & \sout{100.0\%}\phantom{0} \\
MetricAvg
& 85.0\% & 106.9\% \\
MetricVar
& 81.0\% & 97.2\%  \\
MetricCons
& \hspace{-3.5mm} $\star$ 71.4\% & 90.1\% \\
Diversity
& 76.8\% & \hspace{-3.5mm} $\star$ 78.7\% \\
Diff.$\times$Disc.
& 87.2\% &  91.5\% \\
\bottomrule
\end{tabular}

\vspace{-2mm}
\caption{Proportion of data needed to reach the same evaluation result for WMT23 (soft pairwise accuracy) as random subset selection.
Averaged across budgets from 
\Cref{fig:main_outputbased,fig:main_sourcebased}.
}
\label{tab:35-evaluation_parity}

\vspace{-2mm}
\end{table}

\begin{figure}[t]
\centering

\begin{minipage}{\linewidth}
\small
\legendShort{000000}{Random}{}
\legendShort{bc272d}{MetricAvg}{}
\legendShort{50ad9f}{MetricVar}{}\\
\legendShort{0000a2}{MetricCons}{}
\legendShort{e9c716}{Diversity}{}
\legendShort{8c6e96}{Diff.×Disc.}{}
\end{minipage}

\includegraphics[width=\linewidth]{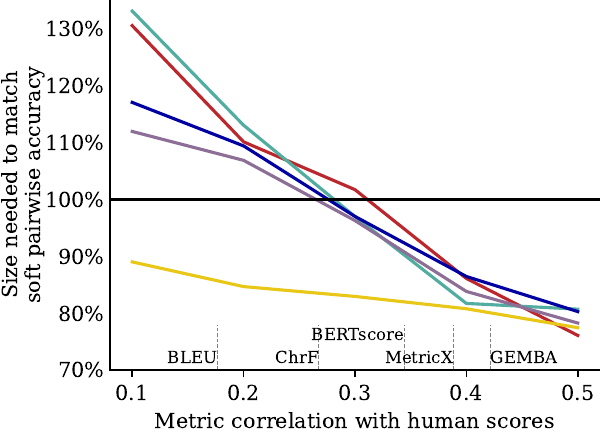}

\vspace{-2mm}
\caption{
Proportion of data needed to reach the same evaluation result for WMT23 (soft pairwise accuracy) with respect to the automated metric used for informing the selection.
The automated metric correlation quality (x-axis) is measured as item-level Pearson correlation against human scores.
Each value is a separate dataset and metric, averaged in bins 0.1, 0.2, 0.3, 0.4, 0.5.
Even though the diversity method is invariant to the metrics, it is affected by the particular datasets characteristics (e.g. quality of human annotations), which differ across bins.
}
\label{fig:16-metric_quality_performance}

\vspace{-3mm}
\end{figure}

\subsection{Importance of a good automated metric}
\label{sec:metric_quality_performance}

In our main results for machine translation (\Cref{fig:main_outputbased}), we used MetricX-23, which is one of the best automated machine translation metrics \citep{freitag-etal-2023-results}. \textit{What happens with a weaker metric?}
In \Cref{fig:16-metric_quality_performance}, we show the relationship between evaluation subset selection performance and quality of the metric, as measured by item-level correlations with human evaluations.
We show results across the 25 metrics available in the WMT metrics shared task \citep{freitag-etal-2021-results,freitag-etal-2022-results,freitag-etal-2023-results}.

MetricX-23 is a pre-trained supervised metric that requires human evaluation data for training.
BERTscore \citep{zhang2020bertscoreevaluatingtextgeneration} is also a pre-trained metric, based on a language model, but is not fine-tuned on human data.
BLEU \citep{papineni-etal-2002-bleu} and chrF \citep{popovic-2015-chrf} are string matching metrics that do not need pre-training or data.
Lastly, GEMBA \citep{kocmi-federmann-2023-gemba} is an LLM-as-a-Judge approach to an automated machine translation metric based on GPT-4.

The string matching metrics (BLEU, chrF) do not provide enough signal for robust evaluation subset selection (worse or on par with random).
However, pre-trained but not supervised metrics (BERTScore) are sufficient because they correlate well enough with evaluations.
Generally, the higher the quality of the automated metric, the better the evaluation subset quality.
However, combining multiple automated metrics into a single item utility does not provide improvements, as shown in \Cref{sec:other_methods}.
Probably, this is because a simple combination of automated metrics rarely leads to higher correlations with human judgments.


\begin{figure}[t]
\centering
\begin{minipage}{\linewidth}
\input{img/30-summeval_spa_legend.tex}
\end{minipage}

\includegraphics[width=\linewidth]{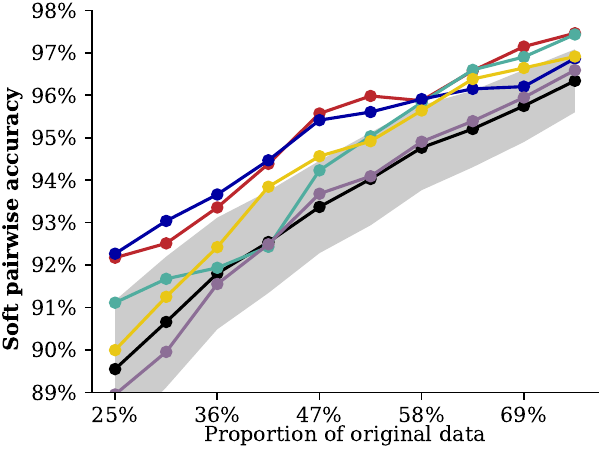}

\caption{
Main \textbf{output-based results for SummEval} averaged across relevance, coherence, consistency, fluency, and their sum with soft pairwise accuracy.
The utility predictors are metrics moments (MetricAvg, MetricVar), consistency (MetricCons) and Item Response Theory (DiffDisc) based on G-Eval, and output diversity.
Numbers in legend show averages across all points.
We also show 90\% t-distribution confidence intervals for the Random Selector from 100 runs. Numbers in the legend show average soft pairwise accuracy across all data proportions.
}
\label{fig:main_summeval}
\end{figure}

\begin{table}[t]
\centering
\small
\begin{tabular}{lcc}
\toprule
\bf Method & \bf Human & \bf G-Eval   \\
\midrule 
\sout{Random} & \sout{100.0\%}\phantom{0} & \sout{100.0\%}\phantom{0} \\
MetricAvg  & 73.7\% & 84.0\% \\
MetricVar  & 82.8\% & 84.7\% \\
MetricCons & 70.6\% & 87.5\% \\
Diversity  & 88.7\% & 89.2\% \\
DiffDisc   & 99.5\% & 78.6\% \\
\bottomrule
\end{tabular}

\caption{
Proportion of data needed to reach the same evaluation result on SummEval (soft pairwise accuracy) as random subset selection with respect to human on LLM evaluation. Averaged across \Cref{fig:main_summeval} budgets.
}
\label{tab:30-summeval_parity}

\end{table}


\section{Case Study 2: Summarization}
\label{sec:results_summ}

We now show the applicability of our methods to another natural language generation task, summarization.
The summarization task is more open-ended than machine translation, making the evaluation even more difficult.
In this section, we explore budget-efficient subset selection for both human evaluation and cheaper, yet still not free, large language model-based evaluation.

\paragraph{Setup.}

We use SummEval \citep{fabbri-etal-2021-summeval} which contains 17 human-evaluated models on 100 items.
The human evaluation of each model output, in contrast to machine translation, is not a single scalar for overall quality.
Instead, humans evaluate the output in four dimensions: \emph{relevance}, \emph{coherence}, \emph{consistency}, and \emph{fluency}.
We use the subset selected for evaluation on each quality dimension independently and then aggregate the final results.
We choose G-Eval \citep{liu-etal-2023-g}, an LLM-based evaluator, as it has a high correlation between metric predictions and human scores.

\paragraph{Results.}

The results of output-based data selection for summarization are shown in \Cref{fig:main_summeval}.
We do not include source-based selection because that requires training data, which are not available for a dataset this small.
In most cases, the methods outperform the random selection.
Similarly to machine translation, the metric average is not consistently very good.
Metric variance, consistency, and diversity are again much stronger.
The results in \Cref{tab:30-summeval_parity} show that we only need $\sim70\%$ of the budget to obtain the same evaluation result as random selection when using the metric consistency selector.
We show the results across the individual evaluation dimensions (relevance, coherence, consistency, and fluency) in Appendix \Cref{tab:48-everything_results_spa}.

Even when using the LLM-based metric for final evaluation, it can become expensive on a larger scale.
Budget-efficient subset selection can also make even this type of evaluation more economical.
The results in the right column of \Cref{tab:30-summeval_parity} show that we can again select only $\sim80\%$ of the test set to be LLM evaluated to reach the same evaluation result as random selection.

\begin{table*}[t]
\newcommand{\emojiT}{\raisebox{-2mm}{\includegraphics[width=4mm]{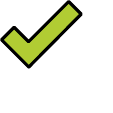}}}
\newcommand{\emojiF}{\raisebox{-2mm}{\includegraphics[width=4mm]{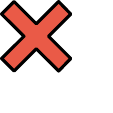}}}
\newcommand{\emojiM}{\raisebox{-1mm}{\includegraphics[width=3mm]{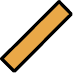}}\,\,}
\centering
\small
\begin{tabular}{llll@{\hspace{1mm}}l@{\hspace{1mm}}l@{\hspace{1mm}}l@{\hspace{1mm}}l@{\hspace{1mm}}}
\toprule
\bf Work &
\bf \makecell[l]{Human-eval.}\hspace{-2mm} &
\bf Scores &
\bf \makecell[l]{Output-based} & 
\bf \makecell[l]{Source-based} &
\bf \makecell[l]{Cost-aware} \\
\midrule
Ours &
\emojiT yes &
\emojiT continuous &
\emojiT yes &
\emojiT yes &
\emojiT yes \\
\citet{rodriguez-etal-2021-evaluation} & 
\emojiF ground-truth &
\emojiF binary &
\emojiT yes & 
\emojiF no &
\emojiF no \\
\citet{ni2024mixeval}&
\emojiF ground-truth &
\emojiF binary &
\emojiF no &
\emojiT yes &
\emojiF no \\
\citet{polo2024tinybenchmarks}& 
\emojiF ground-truth &
\emojiM continuous* & 
\emojiT yes &
\emojiM partly & 
\emojiF no \\
\citet{vivek-etal-2024-anchor}&
\emojiF ground-truth &
\emojiF binary &
\emojiT yes &
\emojiT yes & 
\emojiF no \\
\citet{feng2024sampleefficienthumanevaluationlarge} &
\emojiT yes &
\emojiF pairwise & 
\emojiT yes &
\emojiF no &
\emojiF no \\
\citet{ashury-tahan-etal-2024-label} &
\emojiT yes &
\emojiF pairwise &
\emojiT yes &
\emojiF no &
\emojiF no \\
\citet{ruan2024betterrandomreliablenlg} &
\emojiT yes &
\emojiT continuous &
\emojiM iterative &
\emojiF no &
\emojiF no \\
\citet{li2025autobencher} &
\emojiF no &
\emojiF binary &
\emojiT yet &
\emojiF no &
\emojiF no \\
\bottomrule
\end{tabular}

\caption{Prior work on budget-efficient evaluation subset selection. Ground truth: needs comparison to ground truth, scores: binary or continuous outcomes, or pairwise comparisons, output/source-based: methods for the two subset selection variants, cost-aware: can take evaluation cost into account.}
\label{tab:comparison_related_work}
\end{table*}

\section{Related Work}
\label{sec:related_work}

In this section, we provide context for natural language generation evaluation and an overview of previous works on budget-efficient evaluation.


\paragraph{NLG evaluation.}

Machine translation is one of the most important NLG tasks, and the progress of MT is tracked annually in various WMT evaluations \citep{kocmi2024wmt,wat-2024-1,ahmad-etal-2024-findings-fixed}.
For some NLG tasks, the output is a string in natural language, and its quality can not always be easily assessed by comparing to a ground truth.
This is because there might be many acceptable outputs, such as multiple equally correct translations of a single sentence.
For this reason, robust NLG evaluation eventually relies on human annotators.
However, the human evaluation process is also not straightforward, ranging from assigning a single score \citep{graham-etal-2015-accurate,kocmi-etal-2022-findings} to marking error spans \citep{lommel2014multidimensional,kocmi-etal-2024-error}.
Typically, the output of this process is a single number per input+output, which is used to determine the final model ranking.

At scale, such as during model development, human evaluation is too costly.
For this reason, automated metrics have been developed, starting from text matching approaches \citep{papineni-etal-2002-bleu,popovic-2015-chrf}, progressing to learned metrics \citep{rei-etal-2020-comet,juraska2024metricx24}.
Although learned metrics correlate more strongly with humans \citep{freitag-etal-2022-results}, they also have unexpected problems \citep{zouhar-etal-2024-fine,zouhar-etal-2024-pitfalls,falcao-etal-2024-comet-low}.
Like human annotations, automated metrics also produce a single number for each input+output, evaluating the quality of the NLG model's output. 

Datapoint selection already begins to play a role for automated metric meta-evaluation.
For example, for summarization metrics evaluation, \citet{deutsch-etal-2022-examining} redefine high quality metric to be those being able to distinguish between closely-performing models.
To this end, the metrics performance is not meta-evaluated with correlation across the whole set (even if human scores are available), but on a subset of closely-matched systems.

\paragraph{Budget-efficient evaluation.}
Previous works attempted to find the most informative or diverse evaluation items, primarily by comparison to a ground-truth answer.
In contrast, our goal is to reduce the number of \textit{human} evaluations, while still arriving at the same result, such as model ranking.

To select the most informative items for evaluation, \citet{rodriguez-etal-2021-evaluation} and \citet{polo2024tinybenchmarks} use Item Response Theory (IRT, \citealp{santor1998progress}), a framework for the creation of test sets for students in classrooms.
However, these methods require ground-truth answers to which the model output is compared with a binary outcome.
For natural language generation tasks, there is no single ground-truth answer, and the model output evaluation is a continuous score.

In a more complex version of stratified sampling \citep{saldias-fuentes-etal-2022-toward},
\citet{ni2024mixeval} supersample items to mimic the distribution throughout the test set.
The subset is constructed based on most difficult items, which again requires a comparison to a ground truth and a previous evaluation of some models on the whole test set.
For the evaluation of classification models, \citet{vivek-etal-2024-anchor} find anchor points that describe the outcome of the evaluation on the whole test set.
\citet{feng2024sampleefficienthumanevaluationlarge} choose items that have the least similar model output to be evaluated via pairwise comparisons, which is not applicable to direct model output evaluation.


Many previous works require that at least a few models have already been evaluated on the entire set of items from which we select a subset \citep{rodriguez-etal-2021-evaluation,feng2024sampleefficienthumanevaluationlarge,ruan2024betterrandomreliablenlg}.
This makes the methods not applicable for source-based selection, where either the evaluation set is too large to be processed or where the to-be-evaluated models are not known in advance.

Subset selection through batched active learning \citep{park2022active,mendonca-etal-2021-online,mendonca-etal-2023-onception,ruan2024betterrandomreliablenlg,li2024activeevaluationacquisitionefficient} is another approach using iterative selection.
In practice, human evaluations are usually run all at once or outsourced to a third party with annotators working asynchronously.
Therefore, active learning is possible only when the annotation process is tightly controlled.
In contrast, our item utilities can be simply used to sort the items from most to least informative, and annotators can then stop ad-hoc when the budget is reached.

\citet{zouharkocmi2024esaai} already human-evaluate only a subset of the whole test set by skipping items where automated metrics report no errors.
However, this subset selection is ad hoc and does not offer any control over the subset size.

To our knowledge, no prior work has studied subset selection for evaluation in which the item evaluation cost is taken into account.
See \Cref{tab:comparison_related_work} for a high-level comparison with previous work.

\paragraph{Coresets.}
Discovering a coreset of the data is another potential approach.
A coreset of a dataset is defined as a small subset such that solving a problem on the coreset yields the same result as solving the same problem on the original set \citep{jubran2019introductioncoresetsaccuratecoresets}.
Although appearing like the golden bullet for our problem, it is not applicable for two reasons.
First, we do not know the information (human scores) of the full set that we are trying to reduce.
Second, the coreset algorithms make use of properties of the loss (subset evaluation) function.
Although optimization algorithms exist to, for example, find a subset of vectors with the same mean (approximate mean coreset, \citealp{pmlr-v124-vahidian20a}), to our knowledge, none exist to optimize soft pairwise accuracy or other related ranking evaluation.
Finally, coreset algorithms are often about reducing computational costs in subset construction, which is not the case in our setup.
In \Cref{sec:other_methods} we examine a brute-force imitation of coreset construction and show that due to the first point (lack of human scores at the time of selection), these methods fall short.

\section{Conclusion and Key Takeaways}

We formalize the task of selecting subsets from the test set
with the goal of selecting a subset of the test set for efficient human evaluation.
We explore two common variants of this task: {source-based selection} (no model outputs available), and {output-based selection} (outputs and automated metrics available).
We present several methods based on metric variance, consistency, and diversity; and show that they outperform the dominantly used random selection approach for machine translation (\Cref{sec:results_mt}) and summarization evaluations (\Cref{sec:results_summ}).
However, the simple heuristic of using difficulty estimate by prioritizing items with lowest automated metric scores does not lead to large improvements and can be counterproductive as it prioritizes costly-to-annotate items.
All our methods are implemented in the \href{https://github.com/zouharvi/subset2evaluate}{subset2evaluate} package with pre-trained item utility estimators ready to use for subset selection in machine translation.

\paragraph{Takeaways.}
Based on the analysis in this paper, we offer the following advice to NLG researchers and practitioners for selecting a subset of data for human evaluation.
\begin{itemize}[left=0mm,topsep=0mm,noitemsep]
\item If model outputs and reliable metrics are available, use metric variance or metric consistency.
\item If model outputs are available but a reliable metric is not available, use output diversity.
\item If model outputs are not available but historical data are available, use the diversity\textsuperscript{src} estimation.
\item If only some model outputs are available, use an artificial crowd with metric variance.
\end{itemize}

\paragraph{Limitations.}
A key limitation of our approach is the potential bias of automated metrics and artificial crowd selection.
Automated metrics may already have been used to inform evaluation procedures, although they are misaligned with the final human judgments \citep{kocmi2024preliminary,kocmi2024wmt}.
Data selection for human evaluation would make it harder to catch these biases.
To illustrate this issue, consider a model that underperforms on specific types of item that is not detected by the automated metric.
As these items are estimated to not be faulty, they are not selected for human evaluation.
This issue mainly arises when using expected metric averages for item selection, although similar challenges could occur with variance- or discrimination-based methods (see discussion in \Cref{sec:results_mt_sourcebased}).
\Cref{sec:metric_quality_performance} shows that the performance of our methods directly corresponds to the alignment between automated metrics and human judgments.
Thus, our methods would continue to benefit from future improvements in automated metrics.

\paragraph{Ethical considerations.}

We reuse existing data from WMT and SummEval and do not employ our own annotators.
In the context of automatization and job security, our work aims not to substitute human work, but to ensure that the effort is not wasted on annotating less informative items, ultimately making the work more meaningful.













\bibliography{misc/bibliography.bib,misc/anthology.min.bib}
\bibliographystyle{misc/acl_natbib}

\clearpage

\appendix
\begin{table*}[htbp]
\small
\centering
\begin{tabular}{lc}
\toprule
\bf Method & \bf SPA \\
\midrule
Oracle subset (human, N=10) & 94.1\% \\
Oracle subset (human, N=100) & 95.3\% \\
Oracle subset (human, N=1000) & 96.1\% \\
Oracle subset (MetricX, N=10) & 91.6\% \\
Oracle subset (MetricX, N=100) & 91.7\% \\
Oracle subset (MetricX, N=1000) & 91.7\% \\
Greedy subset (human, N=10, S=10) & 97.0\% \\
Greedy subset (human, N=100, S=10) & 97.9\% \\
Greedy subset (MetricX, N=10, S=10) & 92.2\% \\
Greedy subset (MetricX, N=100, S=10) & 91.8\% \\
\cmidrule{1-1}
MetricAvg (human) & 92.5\% \\
MetricVar (human) & 92.6\% \\
MetricCons (human) & 96.2\% \\
\cmidrule{1-1}
Diversity (Unigram) & 92.4\% \\
Diversity (ChrF) & 92.1\% \\
Diversity (BLEU) & 92.9\% \\
\cmidrule{1-1}
Sentinel-DA-src & 92.0\% \\
Sentinel-MQM-src & 92.2\% \\
\bottomrule
\end{tabular}
\begin{tabular}{l>{\hspace{-1mm}}c}
\toprule
\bf Method & \bf SPA \\
\midrule
k-means (sources) & 92.3\% \\
k-means (sources, weighted) & 91.5\% \\
k-means (outputs) & 91.8\% \\
k-means (outputs, weighted) & 89.3\% \\
\cmidrule{1-1}
DiffUse & 92.4\% \\
Better than Random & 91.2\% \\
\cmidrule{1-1}
Item Response Theory (difficulty) & 89.0\% \\
Item Response Theory (discriminability) & 88.7\% \\
Item Response Theory (feasability) & 84.7\% \\
Item Response Theory (Fisher information) & 92.5\% \\
\cmidrule{1-1}
COMET-instant-confidence (high error) & 91.1\% \\
COMET-instant-confidence (low error) & 85.1\% \\
\cmidrule{1-1}
MetricAvg (MetricX) + MetricVar (MetricX) & 93.1\% \\
MetricAvg (MetricX) + MetricAvg (XCOMET) & 92.6\% \\
MetricVar (MetricX) + MetricVar (XCOMET) & 93.5\% \\
MetricCons (MetricX) + MetricVar (MetricX) & 93.2\% \\
MetricCons (MetricX) + Diversity (LM) & 93.1\% \\
\bottomrule \\[-2mm]
\end{tabular}

\vspace{-1mm}
\caption{
Results for other methods described in \Cref{sec:other_methods} on WMT23 machine translation evaluation measured with soft pairwise accuracy.
\textit{MetricX} is MetricX-23 and \textit{XCOMET} is XCOMET-XXL.
}
\label{tab:43-other_methods}

\vspace{-2mm}
\end{table*}

\section{Other Methods}
\label{sec:other_methods}

We now describe and evaluate a collection of subset selection methods which are not included in the main paper either because of their assumptions or low performance, which limits their application.

\subsection{Methods}

For simplicity and comparability across methods that do not support cost-aware subset selection, we assume constant cost of each item in this section.
Thus we assume $\mathrm{Cost}(\mathcal{Y}) = |\mathcal{Y}|$ in the optimization objective from \Cref{eq:problem_formulation}.

\paragraph{Bruteforce subset selection.}
For the bruteforce approach we simply sample $N$ subsets $\hat{\mathcal{Y}}$ of size $|\hat{\mathcal{Y}}| = B$ and pick the one with the highest utility $\mathrm{Utility}(\hat{\mathcal{Y}})$.
Evaluating $\mathrm{Utility}({\mathcal{Y}})$ requires computing soft pairwise accuracy with respect to the full set $\mathcal{X}$ using human scores, to which we do not have access at the time of subset creation.
However, we can either consider the human scores as an oracle, or compute soft pairwise accuracy with respect to the proxy metric, such as MetricX-23.
Sampling only 1 subset corresponds to the random baseline.
Importantly, this optimization does not require the second simplification put forth in \Cref{sec:problem_statement}, namely that items contribute value to the evaluation independently and not jointly.

\paragraph{Greedy subset selection.}
The previous method relies solely on the variance of random selection and taking the maximum (as in \Cref{fig:main_outputbased}).
A natural extension is to start with a small random set and iteratively extend it by examples that increase the soft pairwise accuracy the most.
Specifically, we keep $\hat{\mathcal{Y}}_i$ at each step $i$ and choose the extension with size $S$:

\vspace{-5mm}
\begin{align}
\hat{\mathcal{Y}}_{i+1} = \hat{\mathcal{Y}}_{i+1} \cup \underset{\substack{\mathcal{Y^+}\subseteq\mathcal{X} \setminus \hat{\mathcal{Y}}_i\\ |\mathcal{Y}^+| = S}}{\arg\max} \,\, \mathrm{Utility}(\mathcal{\hat{\mathcal{Y}}_{i} \cup \mathcal{Y}^+})
\end{align}
\vspace{-2mm}

\noindent
We stop iterating steps when $|\hat{\mathcal{Y}}_{i}| = B$.
The bruteforce subset selection is a special case where $S=B$.
The greedy subset selection is also a special case of beam search, with beam size of 1.

\paragraph{Human oracle.}
The existing methods, such as greedy subset selection, MetricAvg, MetricVar, and MetricCons reliy on automated metrics as proxy to human scores.
To see how much are these methods limited by the proxy being imperfect, we replace the metrics with the human scores.
This would make these methods illegible for real subset selection and thus should be treated rather as an oracle.

\paragraph{Item response theory.}
For the item response theory in the main paper, we use the product of difficulty $b_x$ and discriminability $a_x$ as item $x$'s utility.
Now, we include using the difficulty $b_x$, discriminability $a_x$, and feasability $c_x$ alone.
We also include Fisher information, which corresponds to the variance of the prediction (Equation \ref{eq:irt}) with respect to the model ability $\theta_m$.
The $\mathrm{FI}(x, \theta)$ is defined as:
\begin{align}
& \hspace{-1mm} \phantom{=}-\mathbb{E}\left[\frac{\partial^2}{\partial \theta^2} \log \frac{c_x}{1+\mathrm{exp}\left[-a_x (\theta - b_x)\right]} \Big| \theta \right]\\
& \hspace{-1mm} = \phantom{-} \mathbb{E} \left[ \frac{a_x^2 \cdot \mathrm{exp}\left( a_x \cdot (\theta_m + b_x)\right)}{(\mathrm{exp}(a_x b_x) + \mathrm{exp}(a_x \theta_m))^2} \Big| \theta \right]
\end{align}
This formula is a general form of that of \citet{rodriguez-etal-2021-evaluation,lord2008statistical} which assumes $c_x = 1$ and $b_x = 0$.
Because Fisher information is additive, we sum it across individual model abilities to obtain Fisher information content:
\begin{align}
\mathrm{FIC}(x) = \sum_{m \in \mathcal{M}} \mathrm{FI}(x, \theta_m)
\end{align}

\paragraph{Diversity.}
Previously we have compute model output diversity using vector similarity based on multilingual embeddings of MiniLM-12-v2 \citep{reimers-gurevych-2019-sentence}.
However, this similarity can easily be replaced with simpler string-matching metrics, such as chrF \citep{popovic-2015-chrf}, BLEU \citep{papineni-etal-2002-bleu}, or Dice-Sørensen coefficient between unigrams.
These metrics are averaged across pairwise comparisons between all model output pairs.

\paragraph{Sentinel \citep{perrella-etal-2024-guardians}.}
The Sentinel metric has been proposed for the WMT Metrics Shared task to measure how much quality estimation relies on the source item difficulty rather than estimating the output quality.
It is akin to MetricAvg\textsuperscript{src} and also based on XLM-RoBERTa \citep{conneau-etal-2020-unsupervised}, though without target-side aggregation (averaging).

\paragraph{Clustering.}
A common approach to benchmark creation is to maximize the diversity of the (source) items, such that the new test set covers diverse topics.
This is not yet directly captured by any of the other methods.
To directly optimize this, we embed each item into a vector, run k-means clustering with $k=B$, and for each each of the cluster choose the item to represent it whose embedding is closest to the cluster center.
This ensures that items are chosen far away from each other in the vector space.
The embeddings can be based either on the source texts or the model outputs.

We extend this method further by making an assumption that examples close to each other in the vector space will lead to the same evaluation outcomes.
Based on this, we pretend that all items in the cluster produce the same human scores as the selected item.
In practice, this is done by assigning weight to each selected item based on the size of the cluster it represents.

\paragraph{DiffUse \citep{ashury-tahan-etal-2024-label}.}
The DiffUse method is similar to clustering, but on the space of differences between model representations.
For each item, we compute the distances between each pair.
This then becomes a new vector with the size of $|{\mathcal{M} \choose 2}|$ per each item.
Then, these vectors are merged using hiearchical clustering to fit the predefined budget $B$.
For the embeddings, we use the same multilingual encoder MiniLM-12-v2 \citep{reimers-gurevych-2019-sentence}.

\paragraph{Mixture of methods.}
Finally, we explore mixing methods together.
Most of the methods score items independently to produce utilities.
Our mixing approach is simple:
For each considered utility function, rank the items based on the scores.
Then, average the ranking with weights across multiple utility functions and select top-$B$.

\paragraph{Metric uncertainty.}
Automated metrics, such as COMET \citep{rei-etal-2020-comet}, can be modified to not only provide the best estimate of the translation quality but also an estimate on its own error in the prediction.
In the case of COMET-instant-confidence \citep{zouhar2025earlyexitinstantconfidencetranslation}, the additional prediction is expected absolute error from the true human score, which relates to epistemic uncertainty.
Items with high uncertainty might be those difficulty to annotate and thus also lead to annotation noise, which we want to avoid.
At the same time, the reason why these items are difficult to annotate for a metric might be because in these items the model outputs are most unclear or contested.
In this method, we thus try to sample items with either high or low epistemic uncertainties of the automated metric.

\subsection{Results}

To test the additional methods, we again use the WMT23 dataset and average over languages and data proportions, such that the resulting numbers in \Cref{tab:43-other_methods} are comparable to that in the main paper.

As expected, using human scores when directly optimizing for the best subset leads to outstanding results.
However, this is largely dependent on luck in the random subset generation and taking the maximum.
Instead, building this subset greedily is more efficient and also holds more promise.

Unfortunately, replicating the same approach with automated metrics does not show promise above random.
The best subset as meta-evaluated with respect to an automated metric is not the best subset as meta-evaluated with respect to human scores.
While MetricAvg and MetricVar do not benefit more from using human scores as opposed to the automated metric proxy (see \Cref{fig:main_outputbased} for comparison), MetricCons performs much better than random and the rest of the previously discussed methods.
This means, that further improvements in automated metrics will likely make MetricCons even more useful.

Using k-means clustering for subset selection is both a remedy to potential loss of diversity in the subset, and a well-performing method.
The basic variant based on embedding just the sources and without any reweighting performs the best.
This variant also works as a source-based selector and does not require the knowledge of model outputs, such as DiffUse with comparable performance.
Similarly, even when allowing for immediate feedback with active sampling, the Better-than-random does not perform well in this setup.

Using different pairwise similarity metrics for Diversity does not improve over using embeddings.
Selecting by epistemic uncertainty with COMET-instant-confidence also does not improve over other methods, though items with high uncertainty seem to be prefferable for evaluation to those with low uncertainty.
The Sentinel metrics perform close to the output-based MetricAvg, which its imitates.

Lastly, combining multiple methods together (either same method with different metrics or two methods) using ranking averaging shows only diminishing improvements, which rarely surpasses the individual method's performance.

\begin{table*}[htbp]
\small
\centering
\begin{tabular}{l c c c c c c} \\ \toprule
\bf Meta-evaluation & \bf Random & \bf MetricAvg & \bf MetricVar & \bf MetricCons & \bf Diversity & \bf DiffDisc\\ 
 \midrule 
Soft pairwise accuracy & 91.6\% & 92.6\% & 93.0\% & \bf 93.4\% & 93.0\% & 92.6\%\\ 
Pairwise accuracy & 92.7\% & 93.4\% & 93.4\% & \bf 94.3\% & 93.3\% & 93.2\%\\ 
Pearson correlation & 96.7\% & 96.5\% & 96.5\% & 97.5\% & \bf 97.5\% & 96.9\%\\ 
Spearman correlation & 94.5\% & 95.0\% & 95.1\% & \bf 96.1\% & 95.3\% & 95.1\%\\ 
Kendall\textsubscript{b} correlation & 85.3\% & 86.9\% & 86.9\% & \bf 88.6\% & 86.7\% & 86.3\%\\ 
Top-1 match & 94.6\% & 95.0\% & 95.1\% & \bf 96.4\% & 95.4\% & 96.2\%\\ 
Cluster count & 2.62 & 3.47 & 3.54 & \bf 3.68 & 3.23 & 3.33\\ 
Top-1 cluster match & 59.5\% & \bf 65.6\% & 60.2\% & 64.2\% & 63.3\% & 65.1\%\\ 
Mean average error $\downarrow$ & \bf 0.008 & 0.055 & 0.046 & 0.031 & 0.036 & 0.028\\ 
Mean root squared error $\downarrow$ & \bf 0.010 & 0.062 & 0.059 & 0.039 & 0.041 & 0.034\\ 
\bottomrule \end{tabular}

\vspace{-2mm}
\caption{Output-based subset selection methods on WMT23 meta-evaluated based on \Cref{sec:other_measures}.}
\label{tab:49-other_meta_evaluation}

\vspace{-3mm}
\end{table*}

\section{Other Meta-Evaluations}
\label{sec:other_measures}

There are many ways to measure the usefulness of a subset for evaluation, depending on the goal, which we investigate in this section.

\subsection{Meta-Evaluations}

The meta-evaluation measures how close some evaluation outcome of $\mathcal{Y} \subseteq \mathcal{X}$ is to that of $\mathcal{X}$.

\paragraph{(Soft) pairwise accuracy.}

In the main paper, we use soft pairwise accuracy, which combines the correctness of model ranking on $\mathcal{Y}$ with significance of that ranking.
For each two models $m_1, m_2 \in \mathcal{M}$ we compute the $p$-value of the hypothesis that the averages scores of $m_1$ is higher than to that of $m_2$ (using a paired permutation test, N=1000, \citealp{fisher1935design,goodpermutation}).
These hypotheses are tested within $\mathcal{Y}$ and $\mathcal{X}$, denoted as $p_{m_1>m_2}^\mathcal{Y}$ and $p_{m_1>m_2}^\mathcal{X}$.
This meta-evaluation captures the average similarity in confidences of pairwise comparisons:
\begin{align}
& \hspace{-2mm} \mathrm{SPA}(\mathcal{Y}, \mathcal{X}) = \\
& \hspace{5mm} \frac{1}{|{\mathcal{M} \choose 2}|} \sum_{m_1, m_2 \in {\mathcal{M} \choose 2}} 1 - |p^\mathcal{X}_{m_1>m_2} - p^\mathcal{Y}_{m_1>m_2} | \nonumber
\end{align}
The soft pairwise accuracy has been proposed by \citet{thompson-etal-2024-improving} and simple pairwise accuracy is a special case where where the $p$-values are 0/1 based on if $m_1 > m_2$.

\paragraph{Correlations.}
Possibly the simplest evaluation is model ranking, which can be compared between the subset $\mathcal{Y}$ and the full set $\mathcal{X}$.
This comparison can be done with a correlation, such as Pearson, Spearman, or Kendall variant $b$.
The Pearson correlation stands out by taking the scale into consideration, which makes it sensitive to outliers.

\paragraph{Top-1 match.}
In some cases, the purpose of the evaluation is to find the best model.
For this, we measure how often the top model in $\mathcal{Y}$ to be the same as that of $\mathcal{X}$.
Similar to \citet{deutsch-etal-2022-examining}, this omits many of the evaluated examples to align the meta-evaluation with the practical goal of selecting the best model.

\paragraph{Model average error.}

In cases where we are not interested in the model ranking, but accurate absolute scores, we can measure the mean absolute error and (root) mean squared error between the model average on $\mathcal{Y}$ and the model average on $\mathcal{X}$.

\paragraph{Cluster count.}
In the General WMT Shared Task \citep{kocmi-etal-2024-findings}, greedily compute clusters based on model ranking, as in \Cref{alg:cluster_count}.
Because all models in one cluster are statistically better than the models in the next cluster, the goal is to have evaluation that leads to as many cluster numbers as possible.

\begin{figure}[htbp]
\begin{minipage}{\linewidth}
\hrule \vspace{1mm}
\fontsize{9.5}{10}\selectfont
\textbf{Inputs}: Models $\mathcal{M}$
\hfill
\textbf{Output}: Number of clusters $|C|$
\vspace{0.5mm}

Add a system to the same cluster if it is not  statistically distinguishable from the previous cluster.
\hrule
\vspace{0.5mm}
\fontsize{9.5}{9}\selectfont
\begin{algorithmic}[1]
\State $\mathcal{M} \gets \textsc{Sort}(\mathcal{M}, \lambda m: -\textsc{Avg}(m))$
\State $C \gets \langle \, \langle S_{0} \rangle \, \rangle$
\State \textbf{for}  $m \in \mathcal{M}_{>1}$
\State \hspace{0.5em} \textbf{if} {$\textsc{Wilcoxon}(C_{-1,-1}, m) < 0.05$}
\State \hspace{1em} $C.\textsc{append}(\langle m \rangle)$
\State \hspace{0.5em} \textbf{else} 
\State \hspace{1em} $C_{-1}.\textsc{append}(m)$
\State \Return $|C|$
\end{algorithmic}
\vspace{0.5mm}
\hrule
\vspace{1.5mm}
\captionof{algorithmCaption}{Computation of number of clusters given an evaluated set of items.}
\label{alg:cluster_count}
\end{minipage}

\vspace{-4mm}
\end{figure}

\paragraph{Top-1 cluster match.}
Evaluation with the top-1 match does not distinguishing between cases where there is only a small difference between the top models and cases where the top-1 model is the sole winner.
To remedy this, we can compare the similarity between the top-1 clusters (computed as in \Cref{alg:cluster_count}) based on $\mathcal{Y}$ and $\mathcal{X}$.
For this, we use the Sørensen–Dice coefficient:
$\frac{2\times |C_1^\mathcal{Y} \cap C_1^\mathcal{X}| }{|C_1^\mathcal{Y}|+|C_1^\mathcal{X}|}$.

\paragraph{Active sampling.}
\citet{ruan2024betterrandomreliablenlg} proposes a constrained active sampling framework that iteratively selects examples for inclusion in $\mathcal{Y}$.
This method combines a learned model for quality prediction, systematic sampling to ensure diversity across quality levels, and a constrained controller to reduce redundancy. 
However, due to its active sampling nature, it requires immediate feedback of the human score upon selecting an item, which is not compatible with the selection of the whole subset at once prior to human evaluation.

\subsection{Results}

The results are shown in \Cref{tab:49-other_meta_evaluation} for output-based subset selection methods of the main paper.

The biggest gains over the random selection can be seen for the cluster count, with up to +1 cluster when using MetricCons.
This improvement is meaningful, because in the context of a shared task, having an extra significance clusters allows for more statistically justified claims.
Further, even though the improvements on correlation-based meta-evaluations over random are not by a large margin, they are consistent.
The only exceptions are the differences from the model averages.
However, this is not unexpected and rarely the goal of the meta-evaluation.
Central limit theorem shows that the mean of observed independent random variables (item scores) converges to the true mean (model score average) at least as fast as $x^{-1/2}$ converges to 0.
Sampling not randomly, such as by difficulty (MetricAvg), creates a bias in the average of the random variable, which results in higher error from the mean on the whole set.

\section{Implementation details}
\label{sec:implementation}

\paragraph{Item response theory.}
The item response theory models usually take the form of a logistic regression and are optimized with stochastic variational inference \citep{wu2020variationalitemresponsetheory,rodriguez-etal-2021-evaluation}, which explicitly take into account the distributional priors and dependencies between the latent variables.
The priors for item difficulty ($b$), item discriminability ($a$), and model ability ($\theta$) is the normal distribution and we implement the model in \texttt{py-irt} \citep{py-irt}.

\paragraph{PreCOMET.}
For item utility distillation we use PreCOMET, which is losely based on COMET \citep{rei-etal-2020-comet}.
This model starts as XLM-RoBERTa \citep{conneau-etal-2020-unsupervised}, with attached regressor head (multi-layer perceptron, 768 $\times$ 2048 $\times$ 1024 $\times$ 1), and optimized with Adam (lr=1.5$\times$10$^{-5}$) with weight decay (0.95).
The model is trained for 5 epochs with the effective batch size of 128 on training data of WMT before 2023 (50k sources).
The item utilities are computed with respect to human scores and not automated metrics, since during training human scores are available.

\paragraph{Output diversity.}
We compute the diversity in model outputs using inner products of embeddings in \Cref{eq:diversity} based on multilingual MiniLM-12-v2 \citep{reimers-gurevych-2019-sentence}.

\begin{table*}[htbp]
\fontsize{9.5}{9.5}\selectfont
\centering
\begin{tabular}{llccccccccc}
\toprule
\multicolumn{2}{l}{\bf Dataset} &  \bf \hspace{-4mm}\#Models\hspace{-4mm} & \bf \#Items\hspace{-4mm} & 
\bf Random & 
\bf MetricAvg & 
\bf MetricVar & 
\bf MetricCons & 
\bf Diversity & 
\bf DiffDisc & 
\\\midrule
\parbox[t]{2mm}{\multirow{10}{*}{\rotatebox[origin=c]{90}{WMT24}}}
& \tto{Cs}{Uk} &  11 & 1954 &
92.5\%  & 
93.7\%  & 
\textbf{96.0\%}  & 
92.1\%  & 
93.3\%  & 
93.9\%  & 
\\[0.5em]
& \tto{En}{Cs} &  15 & 571 &
90.3\%  & 
86.5\%  & 
87.0\%  & 
89.6\%  & 
\textbf{92.1\%}  & 
89.0\%  & 
\\
& \tto{En}{Es} &  13 & 634 &
90.7\%  & 
91.0\%  & 
90.5\%  & 
90.5\%  & 
90.1\%  & 
\textbf{91.2\%}  & 
\\
& \tto{En}{Hi} &  10 & 634 &
89.6\%  & 
86.1\%  & 
87.8\%  & 
\textbf{92.5\%}  & 
90.3\%  & 
91.7\%  & 
\\
& \tto{En}{Is} &  10 & 634 &
95.8\%  & 
96.1\%  & 
95.3\%  & 
92.2\%  & 
\textbf{96.4\%}  & 
94.3\%  & 
\\
& \tto{En}{Ja} &  12 & 634 &
85.1\%  & 
82.0\%  & 
\textbf{85.5\%}  & 
84.5\%  & 
85.3\%  & 
84.7\%  & 
\\
& \tto{En}{Ru} &  13 & 634 &
91.1\%  & 
90.4\%  & 
91.1\%  & 
92.7\%  & 
\textbf{92.8\%}  & 
92.1\%  & 
\\
& \tto{En}{Uk} &  10 & 634 &
90.7\%  & 
\textbf{93.8\%}  & 
88.0\%  & 
92.4\%  & 
93.4\%  & 
91.6\%  & 
\\
& \tto{En}{Zh} &  12 & 634 &
\textbf{87.9\%}  & 
87.3\%  & 
86.7\%  & 
86.8\%  & 
84.3\%  & 
86.7\%  & 
\\
& \tto{Ja}{Zh} &  14 & 559 &
93.0\%  & 
95.6\%  & 
\textbf{96.4\%}  & 
94.0\%  & 
94.3\%  & 
94.0\%  & 
\\
\parbox[t]{2mm}{\multirow{9}{*}{\rotatebox[origin=c]{90}{WMT23}}}
& \tto{Cs}{Uk} &  13 & 1009 &
90.0\%  & 
89.1\%  & 
90.4\%  & 
\textbf{94.4\%}  & 
92.5\%  & 
91.6\%  & 
\\[0.5em]
& \tto{De}{En} &  13 & 509 &
91.0\%  & 
91.5\%  & 
91.4\%  & 
92.7\%  & 
\textbf{93.9\%}  & 
92.6\%  & 
\\
& \tto{En}{Cs} &  15 & 1098 &
87.7\%  & 
87.4\%  & 
90.1\%  & 
\textbf{91.8\%}  & 
90.1\%  & 
89.5\%  & 
\\
& \tto{En}{De} &  12 & 549 &
92.1\%  & 
91.8\%  & 
92.1\%  & 
91.9\%  & 
91.7\%  & 
\textbf{92.3\%}  & 
\\
& \tto{En}{Ja} &  16 & 1098 &
93.3\%  & 
\textbf{95.2\%}  & 
94.6\%  & 
94.5\%  & 
94.4\%  & 
94.4\%  & 
\\
& \tto{En}{Zh} &  15 & 1098 &
92.6\%  & 
94.5\%  & 
93.8\%  & 
94.7\%  & 
\textbf{94.9\%}  & 
94.8\%  & 
\\
& \tto{He}{En} &  12 & 820 &
92.9\%  & 
94.9\%  & 
\textbf{96.4\%}  & 
95.6\%  & 
92.8\%  & 
94.1\%  & 
\\
& \tto{Ja}{En} &  17 & 1120 &
92.9\%  & 
\textbf{95.8\%}  & 
95.1\%  & 
92.2\%  & 
95.2\%  & 
94.2\%  & 
\\
& \tto{Zh}{En} &  15 & 884 &
91.4\%  & 
93.3\%  & 
\textbf{94.4\%}  & 
90.3\%  & 
91.1\%  & 
91.0\%  & 
\\
\parbox[t]{2mm}{\multirow{14}{*}{\rotatebox[origin=c]{90}{WMT22}}}
& \tto{Cs}{En} &  11 & 561 &
\textbf{84.7\%}  & 
73.9\%  & 
78.4\%  & 
83.0\%  & 
77.9\%  & 
84.6\%  & 
\\[0.5em]
& \tto{Cs}{Uk} &  10 & 819 &
90.6\%  & 
87.4\%  & 
87.4\%  & 
91.5\%  & 
88.5\%  & 
\textbf{93.1\%}  & 
\\
& \tto{De}{En} &  9 & 601 &
\textbf{80.2\%}  & 
76.7\%  & 
76.4\%  & 
74.9\%  & 
78.5\%  & 
75.6\%  & 
\\
& \tto{En}{Cs} &  10 & 1205 &
89.8\%  & 
90.1\%  & 
\textbf{95.7\%}  & 
92.1\%  & 
89.9\%  & 
90.1\%  & 
\\
& \tto{En}{De} &  14 & 1315 &
91.8\%  & 
\textbf{94.5\%}  & 
\textbf{94.5\%}  & 
90.5\%  & 
92.8\%  & 
92.4\%  & 
\\
& \tto{En}{Hr} &  8 & 1107 &
92.6\%  & 
94.2\%  & 
95.2\%  & 
\textbf{96.8\%}  & 
92.6\%  & 
94.0\%  & 
\\
& \tto{En}{Ja} &  13 & 1181 &
85.9\%  & 
83.2\%  & 
87.2\%  & 
86.9\%  & 
85.0\%  & 
\textbf{87.3\%}  & 
\\
& \tto{En}{Ru} &  15 & 1315 &
95.0\%  & 
95.9\%  & 
95.1\%  & 
95.2\%  & 
94.8\%  & 
\textbf{96.3\%}  & 
\\
& \tto{En}{Uk} &  8 & 1209 &
91.5\%  & 
92.1\%  & 
\textbf{95.3\%}  & 
94.9\%  & 
90.4\%  & 
93.1\%  & 
\\
& \tto{En}{Zh} &  12 & 1181 &
84.1\%  & 
83.1\%  & 
\textbf{89.8\%}  & 
80.4\%  & 
81.0\%  & 
88.9\%  & 
\\
& \tto{Ru}{En} &  10 & 1019 &
85.6\%  & 
86.5\%  & 
86.7\%  & 
83.4\%  & 
82.1\%  & 
\textbf{87.3\%}  & 
\\
& \tto{Sah}{Ru} &  2 & 1023 &
\textbf{100\%\phantom{.}}  & 
\textbf{100\%\phantom{.}}  & 
\textbf{100\%\phantom{.}}  & 
\textbf{100\%\phantom{.}}  & 
\textbf{100\%\phantom{.}}  & 
\textbf{100\%\phantom{.}}  & 
\\
& \tto{Uk}{En} &  9 & 856 &
86.7\%  & 
87.2\%  & 
87.8\%  & 
\textbf{89.2\%}  & 
82.4\%  & 
84.7\%  & 
\\
& \tto{Zh}{En} &  14 & 1875 &
90.2\%  & 
92.2\%  & 
\textbf{95.2\%}  & 
93.1\%  & 
93.9\%  & 
91.4\%  & 
\\
\parbox[t]{2mm}{\multirow{10}{*}{\rotatebox[origin=c]{90}{WMT21}}}
& \tto{En}{De} &  13 & 529 &
83.8\%  & 
87.7\%  & 
85.7\%  & 
\textbf{90.0\%}  & 
85.0\%  & 
85.4\%  & 
\\[0.5em]
& \tto{En}{Ru} &  14 & 512 &
88.3\%  & 
\textbf{94.8\%}  & 
93.0\%  & 
89.3\%  & 
92.7\%  & 
92.9\%  & 
\\
& \tto{Zh}{En} &  13 & 529 &
83.3\%  & 
83.4\%  & 
77.6\%  & 
80.3\%  & 
82.3\%  & 
\textbf{84.9\%}  & 
\\
& \tto{De}{En} &  19 & 653 &
85.9\%  & 
85.2\%  & 
87.2\%  & 
\textbf{89.4\%}  & 
84.7\%  & 
84.7\%  & 
\\
& \tto{En}{Cs} &  10 & 988 &
97.4\%  & 
96.4\%  & 
98.2\%  & 
\textbf{98.7\%}  & 
97.4\%  & 
98.5\%  & 
\\
& \tto{En}{De} &  13 & 527 &
86.8\%  & 
91.1\%  & 
89.5\%  & 
\textbf{93.5\%}  & 
90.1\%  & 
91.0\%  & 
\\
& \tto{En}{Is} &  11 & 838 &
96.0\%  & 
92.4\%  & 
\textbf{96.9\%}  & 
96.3\%  & 
96.4\%  & 
96.2\%  & 
\\
& \tto{En}{Ja} &  15 & 878 &
94.0\%  & 
89.1\%  & 
94.2\%  & 
\textbf{95.2\%}  & 
88.3\%  & 
93.3\%  & 
\\
& \tto{En}{Ru} &  14 & 527 &
86.7\%  & 
\textbf{93.7\%}  & 
91.6\%  & 
91.4\%  & 
87.9\%  & 
91.2\%  & 
\\
& \tto{Zh}{En} &  13 & 650 &
81.3\%  & 
87.3\%  & 
87.7\%  & 
83.0\%  & 
\textbf{89.4\%}  & 
83.3\%  & 
\\
\parbox[t]{2mm}{\multirow{2}{*}{\rotatebox[origin=c]{90}{WMT20}}}
& \tto{Zh}{En} &  8 & 2000 &
88.4\%  & 
\textbf{94.1\%}  & 
91.8\%  & 
86.4\%  & 
88.1\%  & 
91.3\%  & 
\\[0.5em]
& \tto{En}{De} &  7 & 1418 &
90.3\%  & 
86.3\%  & 
85.1\%  & 
91.0\%  & 
85.0\%  & 
\textbf{92.2\%}  & 
\\
\parbox[t]{2mm}{\multirow{4}{*}{\rotatebox[origin=c]{90}{WMT19\hspace{3mm}}}}
& \tto{Kk}{En} &  11 & 1000 &
89.5\%  & 
86.0\%  & 
\textbf{92.8\%}  & 
92.1\%  & 
84.3\%  & 
92.7\%  & 
\\[0.5em]
& \tto{De}{En} &  16 & 1948 &
89.0\%  & 
88.8\%  & 
\textbf{92.9\%}  & 
90.2\%  & 
89.2\%  & 
90.3\%  & 
\\
& \tto{Gu}{En} &  11 & 1016 &
92.2\%  & 
85.0\%  & 
92.2\%  & 
\textbf{95.4\%}  & 
89.6\%  & 
93.0\%  & 
\\
& \tto{Lt}{En} &  11 & 1000 &
92.6\%  & 
91.8\%  & 
91.6\%  & 
89.6\%  & 
92.0\%  & 
\textbf{93.7\%}  & 
\\
\\[-0.2em]
\parbox[t]{2mm}{\multirow{6}{*}{\rotatebox[origin=c]{90}{SummEval}}}
& Relevance &  11 & 100 &
94.0\%  & 
95.7\%  & 
\textbf{95.9\%}  & 
95.5\%  & 
94.2\%  & 
93.3\%  & 
\\
& Coherence &  11 & 100 &
94.7\%  & 
\textbf{95.6\%}  & 
94.5\%  & 
95.0\%  & 
95.3\%  & 
95.1\%  & 
\\
& Consistency &  11 & 100 &
90.2\%  & 
91.8\%  & 
91.2\%  & 
\textbf{92.5\%}  & 
92.0\%  & 
91.0\%  & 
\\
& Fluency &  11 & 100 &
90.2\%  & 
93.9\%  & 
92.0\%  & 
\textbf{94.1\%}  & 
91.1\%  & 
93.1\%  & 
\\
& Sum &  11 & 100 &
95.5\%  & 
\textbf{96.4\%}  & 
95.8\%  & 
96.2\%  & 
96.0\%  & 
94.5\%  & 
\\
& Mul &  11 & 100 &
95.4\%  & 
\textbf{96.1\%}  & 
94.8\%  & 
\textbf{96.1\%}  & 
95.4\%  & 
93.8\%  & 
\\
\bottomrule\end{tabular}

\caption{Individual subset selection results for machine translation evaluation (WMT) and summarization evaluation (SummEval) measured with soft pairwise accuracy averaged over data proportions from 5\% to 50\% (WMT) and 25\% to 75\% (SummEval). Bold numbers indicate best in evaluation category (correlation or clusters) within the row.
Random is average over 100 runs.
}
\label{tab:48-everything_results_spa}
\end{table*}

\end{document}